\documentclass{article}

\usepackage{arxiv}

\usepackage[utf8]{inputenc} 
\usepackage[T1]{fontenc} 
\usepackage{hyperref}  
\usepackage{url}   
\usepackage{booktabs}  
\usepackage{amsfonts}  
\usepackage{nicefrac}  
\usepackage{microtype}  
\usepackage{graphicx}
\usepackage{caption}
\usepackage{subcaption}
\usepackage{xspace}
\usepackage{geometry}
\usepackage{amsmath,amssymb,amsfonts,dsfont}
\usepackage{breqn}
\usepackage{lmodern}

\usepackage{xcolor}

\title{Learning to Optimise General TSP Instances}
\author{Nasrin Sultana, Jeffrey Chan, A. K. Qin, Tabinda Sarwar}

\begin{document}
\maketitle

\begin{abstract}

The Travelling Salesman Problem (TSP) is a classical combinatorial optimisation problem. Deep learning has been successfully extended to meta-learning, where previous solving efforts assist in learning how to optimise future optimisation instances. In recent years, learning to optimise approaches have shown success in solving TSP problems. However, they focus on one type of TSP problem, namely ones where the points are uniformly distributed in Euclidean spaces and have issues in generalising to other embedding spaces, e.g., spherical distance spaces, and to TSP instances where the points are distributed in a non-uniform manner. An aim of learning to optimise is to train once and solve across a broad spectrum of (TSP) problems.  Although supervised learning approaches have shown to achieve more optimal solutions than unsupervised approaches, they do require the generation of training data and running a solver to obtain solutions to learn from, which can be time-consuming and difficult to find reasonable solutions for harder TSP instances. Hence this paper introduces a new learning-based approach to solve a variety of different and common TSP problems that are trained on easier instances which are faster to train and are easier to obtain better solutions. We name this approach the non-Euclidean TSP network (NETSP-Net). The approach is evaluated on various TSP instances using the benchmark TSPLIB dataset and popular instance generator used in the literature. We performed extensive experiments that indicate our approach generalises across many types of instances and scales to instances that are larger than what was used during training.

\end{abstract}


\section{Introduction}\label{intro}

The Travelling salesman problem (TSP) is prevalent in combinatorial optimisation. It requires searching a set of given nodes, which can represent cities or places, to find a route that passes through each node exactly once and has minimal route/tour length. It has numerous applications in telecommunications, circuit board design, DNA sequencing, transportation and in theoretical computer science \cite{applegate2006traveling}. It is an NP-Hard problem, which means that there are instances for which finding an optimal solution is a time-taking process. Traditional approaches to tackle such hard optimisation problems employ two main strategies: exact algorithms \cite{applegate2006traveling}, and heuristics \cite{christofides1976worst}. Exact algorithms are based around enumeration and can solve TSP optimally, e.g., branch-and-bound or integer programming formulations, but they generally cannot scale to larger instances \cite{applegate2006traveling}. Since finding the optimal solutions might not be feasible for a large number of cities, heuristics employ problem-specific knowledge and carefully engineered approaches and parameters to find near-optimal solutions. Therefore, heuristics are effective algorithms which are often fast, but they may be tailored to a specific problem. Application of current search heuristics to new problems or extending to new instances of a similar problem is difficult and challenging. This challenge motivates the level of generalisation at which optimisation systems operate \cite{burke2emerging}, and that is the underlying motivation behind learning to optimise algorithms to solve optimisation problems. Although existing approaches can find reasonable TSP solutions, they need to be restarted for every instance, even where similar instances have been solved before. Instead, the knowledge generated from solving previous instances can be reused and utilised using learning to optimise mechanism, and this can assist in initialising solutions for heuristic and exact solvers for improved and more efficient solution searching. We next describe some related work in learning to optimise area and the open challenges associated with them.

Deep neural networks (DNNs) have boosted performance in machine translation and image processing (\cite{lecun2015deep}). In a similar way, deep learning architecture can be trained to predict solutions to many combinatorial optimisation problems. This new field is called Learning to Optimise \cite{li2017learning}. Learning to Optimise approaches can be divided into supervised \cite{vinyals2015pointer} and reinforcement learning-based \cite{bello2016neural}, \cite{deudon2018learning}, \cite{kool2018attention}, and have learnt to solve classic problems such as Travelling salesman \cite{vinyals2015pointer}, Knapsack \cite{bello2016neural} and Vehicle Routing problems \cite{nazari2018reinforcement}. In the supervised learning-based approaches, TSP instances and the corresponding ground truth solutions are used to train the DNN \cite{vinyals2015pointer}. The reinforcement learning-based approach has been designed to solve combinatorial optimisation problems where policies are learnt to optimise tour length \cite{bello2016neural},\cite{deudon2018learning},\cite{kool2018attention} and can be used as solvers. Although showing good initial promise, learning to optimise for COP suffers challenges. 
 
\begin{itemize}
\item How to predict a solution for general TSP instances? For instance, TSP instances can be characterised by a) the space the points/cities are embedded in, e.g., Euclidean space and b) the difficulty of finding the optimal solution, which is partly related to how the points/cities are distributed within the space. We believe many of the existing approaches focus on training and testing on TSP instances that are embedded in Euclidean space. Also, they focus on generating training and testing instances that are uniformly distributed within a unit hypercube. This can lead to an inability to model, learn on and generalise to TSP instances that are embedded in non-Euclidean spaces and points (nodes) are not uniformly distributed.
 \item How to generate appropriate training data to be able to learn to solve different instances of TSP adequately? Though it is difficult to generate training data using various TSP instances, however, it is easy to generate Euclidean TSP instances/easy instances that we can use as training data to solve all types of TSP instances and easy to find optimal solutions (for small size $(>50)$) of some Euclidean TSP instances using Concorde \cite{applegate2006concorde}. Therefore, we consider generating Euclidean TSP instances as training data that are easy to generate and find a solution and predict solutions for any types of instances. 
\end{itemize}
The motivation of this paper is to present an approach that can able to solve general (various size and difficulty level) TSP instances, given the greater availability of Euclidean instances considering the above mention challenge. We next describe the various size and difficulty level of TSP instances that never analysed before in learning to optimise area in the next two paragraphs. 

\begin{figure*}
  \begin{subfigure}[b]{0.45\textwidth}
  \centering
  \includegraphics[width=\linewidth, height = 5cm]{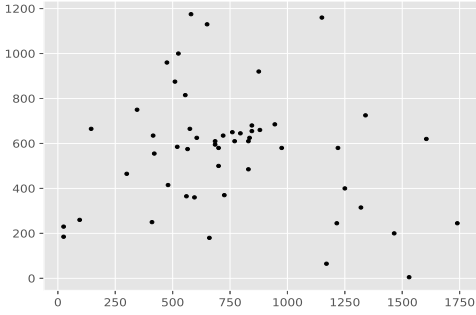}
  \caption{Set of points spread on the space for the problem Berlin52}\label{fig:sfig4}
  \end{subfigure}
  \hfill
  \begin{subfigure}[b]{0.45\textwidth}
  \centering
  \includegraphics[width=\linewidth, height = 5cm]{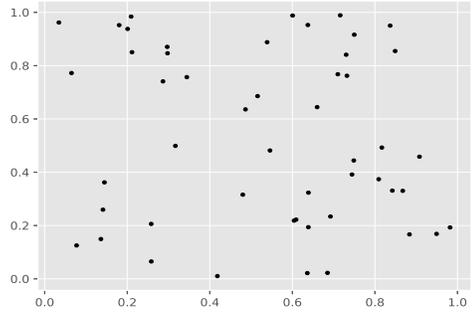}
  \caption{Set of points spread on the space for the problem TSP50 (TSP50 is randomly generated)}
 \label{fig:sfig5}
  \end{subfigure}
  \centering
  \hfill
  \begin{subfigure}[b]{0.45\textwidth}
  \centering
  \includegraphics[width=\linewidth,height = 5cm]{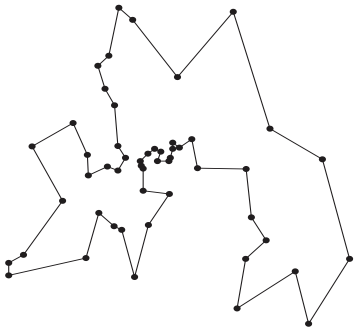}
  \caption{Solution (Berlin52)}\label{fig:sfig6}
  \end{subfigure}
  \hfill
  \begin{subfigure}[b]{0.45\textwidth}
  \centering
  \includegraphics[width=\linewidth,height = 5cm]{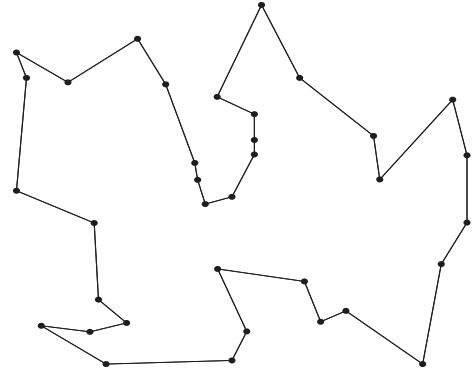}
  \caption{Solution(TSP50)}
 \label{fig:sfig7}
  \end{subfigure}
 \caption{Example of Hard to solve (Berlin52) and easy to solve (TSP50) TSP instances. A graphic representation of the TSP instances and its optimal solutions (solid line); Figure illustrates because of a point on the problem spread on the space alter the solutions. Here, each dot represents a city (points) and the solid line indicates how cities are connected (optimal solution).}\label{fig:hard}
\end{figure*}

Studies have been shown that the main property of measure the difficulty of TSP problem related to how the cities are distributed (and the distance space). In \cite{fischer2005analysis} studied all the TSP instances do not have the same difficulty level such as difficulty/distribution of points (easy, hard and so on) and in terms of size (instance size). We illustrate two examples of TSP instances, for example, hard (Berlin52) and easy(TSP50). In Fig.~\ref{fig:sfig4}, Berlin52 is an example of a hard problem because of the tightly constrained points; it is hard to search solutions that made the problem hard. In Fig.~\ref{fig:sfig5}, TSP50 is an example of an easy problem as the points are loosely constrained, so it is easy to search for solutions for these problems. Hardness generally relates to the distribution of the normalised areas and the distances between the cities. For example, the number of cities, represented as a dot(.) in Fig.~\ref{fig:hard}],  spread in the area covered by the cities of the given problem. This distribution of the cities represent how instances are embedded in the space, and how hard or easy an instance is to solve, as described in \cite{cardenas2018creating}. In \cite{cardenas2018creating}, they classified TSP instances according to their hardness such as easy, and hard to solve instances. In Fig.~\ref{fig:sfig6} and ~\ref{fig:sfig7} is the solution for Fig.~\ref{fig:sfig4} and ~\ref{fig:sfig5} where we illustrate that although instance size is nearly the same for two problems according to distributions of the problems, the solutions also alter on how the number of cities spread in the place. We define hardness indication details in Section~\ref{section:hard}. In this paper, motivation is not to measure the difficulty level of TSP instance, but, we are using the measure to show our model can learn all types of instances (hard and easy) studied in \cite{cardenas2018creating}, which is verified in Section \ref{section:various}.

Another types of TSP instances embedded non-uniform manner with various edge-weight types illustrates in Fig.~\ref{fig:edge}. Edge is a property of new measure of the city to city nearness. Every TSP instances can be classified as their edge-weight types (weight means distances), such as Euclidean distance, Haversine distance\footnote{The edge weights represent the geographic distances (Haversine) between these locations, and TSP points are distributed on a sphere considering the curvature of the Earth.}, pseudo-Euclidean distance\footnote{Pseudo-Euclidean instances, which is Euclidean distance but breakdown of some properties of Euclidean space since the triangular inequality is not satisfied \cite{chatting2018comparison}}. In Fig.~\ref{fig:edge}, we illustrate that every edge-weight type of TSP instance embedded differently in the space, which formulates in Section~\ref{section:3.1}. These instances never analysed before in learning to optimise area. Previous work by the author has evaluated using Euclidean distance TSP instances, and it is anticipated that evaluate other distance types of TSP instances is preferred. Consequently, since we are considering learning general TSP instances so from the TSPLIB library \cite{reinelt1995tsplib95}, we observe various edge-weight types of instances to evaluate our model. We showed our model able to generalise on the different types of common TSP problems that we want one learning algorithm to generalise over, and that existing works have difficulty in generalising. Subsequently, we show the competitiveness of our approach.

In real-world TSP applications, the instances consists of hundreds or thousands of nodes with various structures, and the optimal solution is not able to compute efficiently. We find that the proposed NETSP model generalises well from small-scale TSP problems to larger-scale problems and the various structure of distribution of the problems (cities) and generalisation capacity increases by order of magnitude. Cities may be distributed in a structured or unstructured way, which makes each TSP problem hard \cite{moscato1994analysis}, \cite{cardenas2018creating}. We present a new approach to solve the TSPs embedded in different spaces, and consider the distribution of points, also study different difficulty level of instances \cite{cardenas2018creating}. Subsequently, different edge-weight types \cite{kutkut2001solution} of instances from the TSPLIB benchmark dataset \cite{reinelt1991tsplib} and classify them according to their edge-weight types and hardness. Moreover, we presented that we trained our model with smaller instances but can generalised larger instances. This paper introduces a deep learning approach using a combination of Convolution Neural Network (CNN) \cite{lecun1995convolutional} and Recurrent Neural Network (RNN) \cite{hochreiter1997long}. CNN has been successfully shown to be able to handle different embedding spaces, and as we show in this paper, it is also able to handle TSPs in different spaces. Also, a CNN architecture can discover the inherent micro-patterns within smaller instances that transfer to larger instances and different spaces. In the field of combinatorial optimisation research, we have proposed the Non-Euclidean travelling salesman problem architecture (NETSP-Net) in Fig.~\ref{fig:NETSP}, which aims to learn various TSP instance representations and capture important features (such as local spatial patterns among coordinates to learn about the pattern between two nodes). The objective of this work is not to outperform the current state-of-the-art TSP learning algorithm but to motivate the research in supervised learning  to learn general TSP instances, considering the above difficulties.

The main contributions of our work are as follows:
\begin{itemize}
\item We propose a new architecture, that we call NETSP-Net whose model is applied to non-trivial algorithmic problems involving geometry in various travelling salesman problems.
 \item Our NETSP-Net is applied to different distributions of TSP instances embedded in the Euclidean plane (hard to solve instances). We show that the learned model generalises to various distributions.  
 \item We show that our method can generalise on various TSP instances embedded in various spaces.
\item We show that our supervised model is efficient compared to reinforcement learning and can generalise to larger problems and perform consistently compared to other learning models.
\end{itemize}

\section{Related work}
Before presenting NETSP-Net, we review some background that is related to our model. Over the last few years, much work has been done to tackle combinatorial optimisation problems, particularly from a learning perspective. To begin with, we discuss exact and heuristic solutions, followed by learning algorithms for TSP as well as policy-based methods. 

\subsection {Exact and heuristic solvers}

Combinatorial optimisation problems intensely studied in the operation research field is the travelling salesman problems (TSP). Many algorithms such as exact, heuristics and meta-heuristic can provide the approximate solutions efficiently for larger problems. The exact algorithms (TSP solver Concorde \cite{applegate2006traveling}) can find solution optimally. However, for solving larger problems, the algorithms become slower, and there is an exponential increase in the execution time \cite{lawler1966branch}. In practice, many heuristic approaches are employed including swarm optimisation, simulated annealing, local search, genetic and  greedy algorithms\cite{gendreau2005metaheuristics}. The approximate search heuristic for the symmetric TSP the Lin-Kernighan-Helsgaun heuristic \cite{helsgaun2000effective}, has been shown to solve instances with larger problems optimally. OR-Tools (developed by Google) \cite{ortools} is a solver for vehicle routing problem, tackles the TSP with a combination of local search algorithms and meta-heuristics. Substantially, these approaches need expertise and require extensive problem-specific research.

\subsection {Sequence to sequence learning on Neural Combinatorial Optimisation}

Recent advances in the neural networks include the design of a new model architecture called Pointer Network(PN) \cite{vinyals2015pointer}, inspired by sequence-to-sequence models where the input sequence can determine the output sequence. In sequence-to-sequence models, the overall architecture requires two RNN networks referred to as an encoder-decoder framework. An encoder network receives the input sequence (source information) and encodes the source information into a vector representation. Later, a decoder generates an output sequence from this. In the Sequence-to-Sequence architecture, the encoder takes the input sequence and generates the output sequence based on one vector. For example, \cite{bahdanau2014neural} demonstrates that during decoding steps, using a technique name attention mechanism the decoder can extract important parts of the source information (input sequence) and produces the output sequences using the corresponding information. 

\subsubsection{Neural Combinatorial Optimisation using Supervised Learning Techniques}

Pointer Network(PN) \cite{vinyals2015pointer} learns to solve combinatorial optimisation problems where encoder(RNN) converts the input sequence that is fed to the decoder(RNN). They use attention on the input sequence and trained the model to solve the Euclidean TSP. The PN architecture has been developed to solve TSP,  they trained their network using a supervised method (training date is the randomly generated problem instances with their ground truth optimal (or heuristic) solutions). In \cite{joshi2019efficient} takes a graph as an input and used several graph convolutional layers to extract features from its (graph) nodes and edges. The output of the neural network is an adjacency matrix whose elements represented the probabilities of edges. The edge predictions are then transformed into a feasible tour. Joshi et al. \cite{joshi2019efficient} also trained the model in a supervised manner.

\subsubsection {Neural Combinatorial Optimisation using Reinforcement Learning Techniques}

In recent years, researchers develop reinforcement learning-based frameworks to solve TSP problems \cite{bello2016neural}, \cite{deudon2018learning}, \cite{khalil2017learning}, \cite{kool2018attention}. The proposed model called EAN \cite{deudon2018learning} used Principal Component Analysis (PCA) on the input coordinates. They do not use any recurrent and convolutional layer instead used attention mechanisms \cite{vaswani2017attention}, moreover first time they combine 2OPT local search with improving performance. Model names S2VDQN \cite{khalil2017learning} used a graph embedding structure first in the combinatorial optimisation field. They trained the model using the reinforcement learning deep Q-network (DQN) to develop solutions for graph-based NP-hard problems incrementally. The framework proposed by Kool et al. \cite{kool2018attention}, based on deep reinforcement learning that utilized graph attention layers, demonstrated that high-quality approximate solutions can be achievable to some NP-hard COPs. In \cite{kool2018attention}, construct a solution for TSP by pointing input elements as a sequence. They claim their model is an alternative to the PN \cite{vinyals2015pointer}. Another recent method called (GNN-MCTS), a graph neural network (GNN) \cite{wu2020comprehensive} used Monte Carlo Tree Search (MCTS) to make the decision more reliable by simulating a large number of searches. GNN \cite{wu2020comprehensive} is trained to guide the MCTS, which helps to reduce the complexity of the search space. Also, to avoid stuck in a local optimum MCTS provides a more reliable policy \cite{xing2020graph}. 

In the learning to optimisation field, previous work only considered the TSP instances embedded in a Euclidean plane and randomly generated TSP instances. Our goal is to consider TSP instances according to their edge weight types and distribution. Our model is based on a CNN combined with RNN and attention mechanisms, and this combination significantly improves results. In our work, we have used a CNN, which is more powerful to exploit spatial invariance of all locations (cities). The hardness of TSP instances have been investigated many times \cite{fischer2005analysis},\cite{caldwell2018deep} and solve various edge-weight types \cite{ruffa2007novel} of TSP instances in the operation research field. However, in the learning to optimise field, never address the issue of the hardness of TSP instances and other edge-weight types of instances from the TSPLIB benchmark dataset \cite{reinelt1991tsplib}. We show the benefit of our model in direct comparison to state-of-the-art approaches using randomly generated data, as well as TSPLIB data on various TSP instances.

\section{Background and Problem Formulation}

In this section, we formally define the problem. We first present the TSP problem, then define what learning to optimise means when solving TSP instances.

\subsection{Travelling Salesman Problem (TSP)}

In the TSP problem, we are given a set of  points/cities\footnote{We will use cities and points inter-changeably to mean the same thing.} $C = {C_1,C_2 \cdots C_n}$, embedded in a d-dimensional space, i.e., $C_i \in{\rm \mathbb{R}^d}$ and $d \geq 1$. For routing and navigation, it is typically d = 2, i.e., $C_i \in R^2$, but we stress TSP and our approach can handle problems of any dimensions.  Let, D ($C_i, C_j$), denote a distance measure between cities $C_i$ and $ C_j$, i.e., $D:{\rm \mathbb{R}^d} \times {\rm \mathbb{R}^d \rightarrow \rm \mathbb{R}^{+}} \cup \{0\}$. D can be the Euclidean distance, but can also be other distances which we will shortly introduce. A tour of these cities is a sequence where each city is visited, and only visited once. Let a tour be denoted by $S =S_1, S_2, \cdots, S_n$ where each $S_i \in C$ and $S_i \neq S_j$ $\forall i \neq j$. Then the TSP problem is to find such a tour of cities such that the total travel distance between consecutive pairs of cities in the tour is minimised:
\[
\min_{S}  D(S_{n}, S_{1}) + \sum\limits_{i=1}^{n-1} D(S_{i}, S_{i+1})
\]
\subsection{Popular TSP Distance Measures} \label{section:3.1}
In this section we introduce and formally define popular distance measures used for TSP. These include Euclidean distance, Haversine instances, pseudo-Euclidean instances. 
\subsubsection{EUC2D} 
The Euclidean distance is a popular and commonly used metric to compute distance. It is defined as \cite{kutkut2001solution}:

\[
D(C_i, C_j) = \sqrt{ \sum^{d}_{k=1} (C_i[k] - C_j[k])^2}
\]
where $C_i[k]$ denotes the value of the $k^{th}$ dimension of point/city $C_i$.
\subsubsection{Haversine distance} 

\begin{figure}
  \begin{subfigure}[b]{0.4\columnwidth}
    \includegraphics[width=\linewidth]{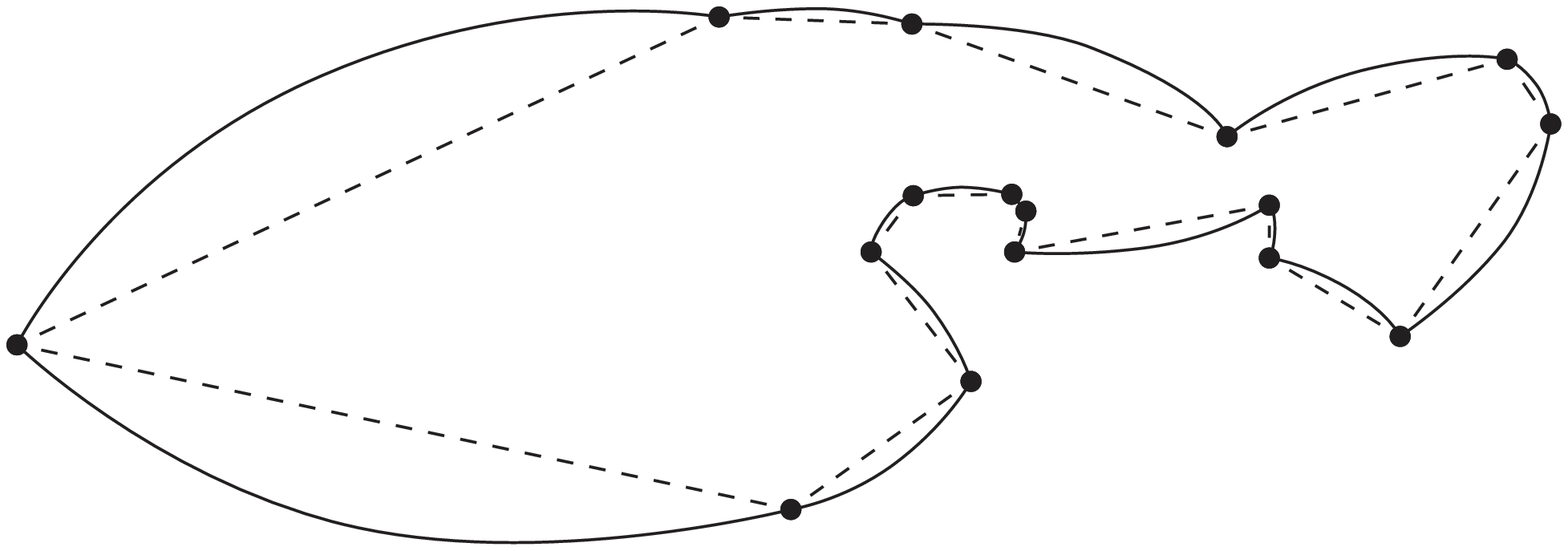}
    \caption{ulysses16}\label{fig:sfig1}
    \label{fig:1}
  \end{subfigure}
  \hfill 
  \begin{subfigure}[b]{0.4\columnwidth}
    \includegraphics[width=\linewidth]{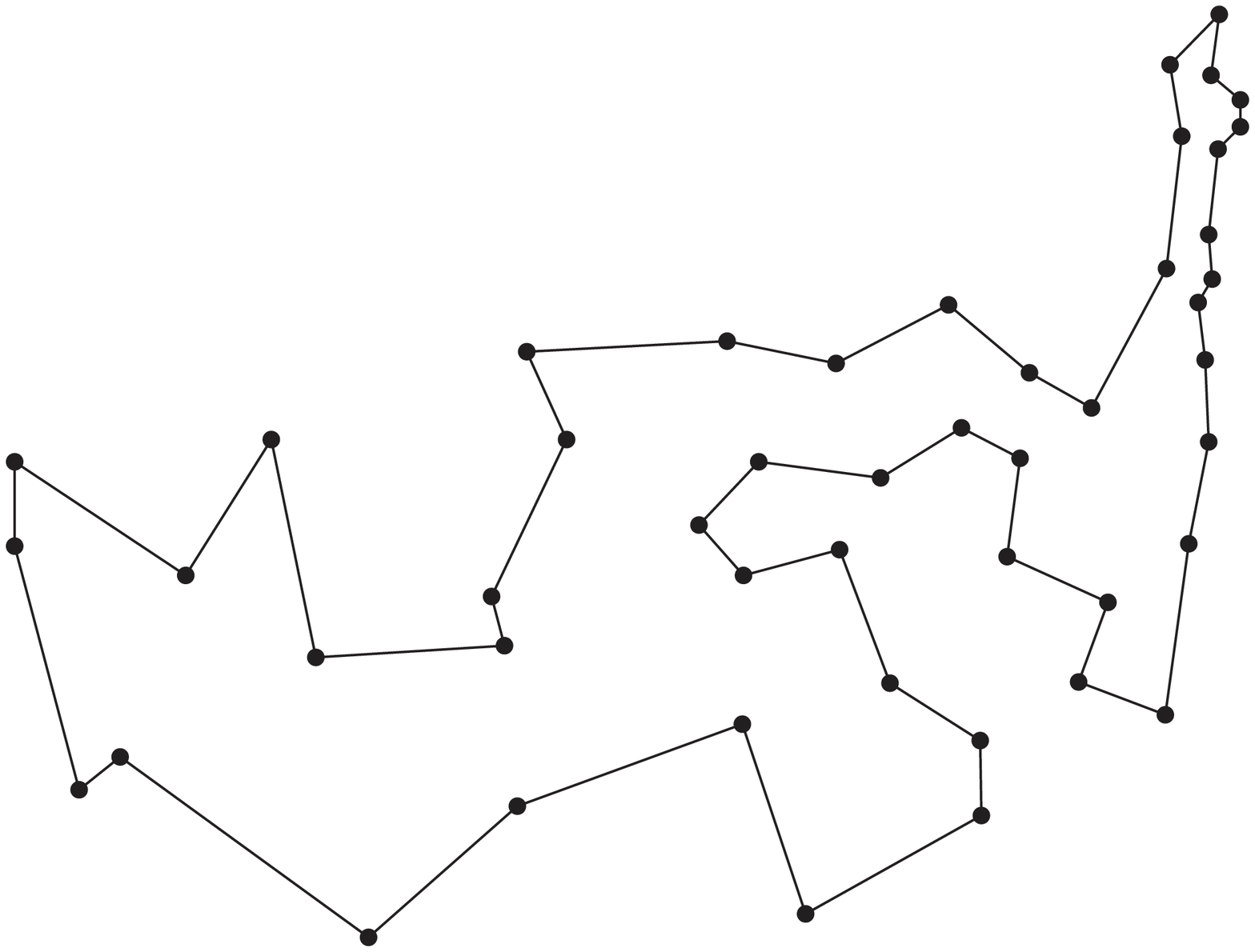}
    \caption{Att48}\label{fig:sfig3}
    \label{fig:2}
  \end{subfigure}
   \caption{Graphical representation of TSP instances in different embedding spaces (dot represents vertices), these can be found as ulysses16.tsp (dashed line represent its optimal solution), att48.tsp in the TSPLIB Benchmark Library \cite{reinelt1991tsplib}}\label{fig:edge}
\end{figure}

The Haversine distance \cite{robusto1957cosine} was introduced to correct and take account of the curvature of the Earth, and a popular distance for navigation and solving TSP where points are distributed on a sphere. The haversine formula determines the great-circle distance between two points on a sphere. It typically defined in terms of latitude and longitude, so consider the case where our cities are described by two dimensions, latitude and longitude,(both in radius). Haversine is important in navigation that relates the sides and angles of spherical triangles. In Fig.~\ref{fig:sfig1}, shows a geographic TSP instance with 16 vertices, name 'ulysses16' \footnote{The distance matrix for this instances was computed using the  Haversine formula (great circle distance)} \cite{prates2019learning}, that is, the shortest round trip. This route is calculated based on direct air distances and can only be travelled by a 'modern Ulysses' using an aircraft. The solid line and arrows indicate the sequence in which Homer's Ulysses supposedly visited the 16 locations. The Haversine distance is thus defined as follows \cite{kutkut2001solution}:

\[
\begin{aligned}
      D(C_i, C_j) &= R \cdot F(C_i, C_j), \\\text{where}\\
F(C_i, C_j) &= 2 A(C_i, C_j) \cdot atan2(\sqrt{A(C_i, C_j)}, \sqrt{1-A(C_i, C_j)}) \\
A(C_i, C_j) &= \sin{(\frac{C_j[1] - C_i[1]}{2})}^2  +  \nonumber \\ & \cos{(C_i[1])} \cos{(C_j[1])} \sin{(\frac{C_j[2] - C_i[2]}{2})}^2
\end{aligned} 
\]
where $C_i[1]$ is latitude, $C_i[2]$ is longitude and $R$ is the radius of the sphere over which the distance is computed, e.g., the radius of Earth.

\subsubsection{Pseudo-Euclidean} 

The instance, ATT48, indicates the edge weight type used is pseudo-Euclidean \cite{reinelt1995tsplib95}. The instance, ATT48, represents the 48 capitals of the adjacent mainland U.S. states and ATT specifies the edge weight type used is pseudo-Euclidean \cite{rennie2017self}. Pseudo-Euclidean \cite{reinelt1995tsplib95} is a Cartesian product of two Euclidean spaces with a specific inner product and the distance. The Pseudo-Euclidean space (PE-space) consists of two orthogonal Euclidean sub-spaces, called the positive space and the negative space. Pseudo-Euclidean is not a metric space as it does satisfy the  triangular inequality \footnote{The pseudo-Euclidean space geometry has a breakdown of some properties of Euclidean space. In particular it is not a metric space \cite{reinelt1995tsplib95}.  For vectors $a$ and $b$ in a real affine space, there exists a definite number, called the scalar product (a,b). In Pseudo-Euclidean space, there are three types of straight lines: Euclidean, ((a, a)>0), pseudo-Euclidean ((a, a)<0) and isotropic ((a, a)=0). Isotropic cone represents the merging of all the isotropic straight lines passing through a point. Furthermore, there are several types of space in a pseudo-Euclidean space, such as Euclidean, pseudo-Euclidean and semi-Euclidean (spaces containing isotropic vectors). This means pseudo-Euclidean space also can be understood as Euclidean space.}. The distance is calculated by finding the difference between co-ordinates of the cities and summing the square of these differences, dividing by ten and calculating the square root. It is defined as \cite{chatting2018comparison}:
\[
D(C_i, C_j) = \sqrt{ \sum^{d}_{k=1} \frac{(C_i[k] - C_j[k])^2}{10}}
\]
where $C_i[k]$ denotes the value of the $k^{th}$ dimension of point/city $C_i$.

In Fig.~\ref{fig:sfig3}, shows a TSP (pseudo-Euclidean space) instance with 48 vertices, name 'att48' and edge represent how cities are connected. So, we motivate to use Att48 instance and haversine instance, which can be considered as non-Euclidean TSP instances, which is different from Euclidean distance instances.
\begin{figure}
 \centering
 \includegraphics[width=0.80\textwidth]{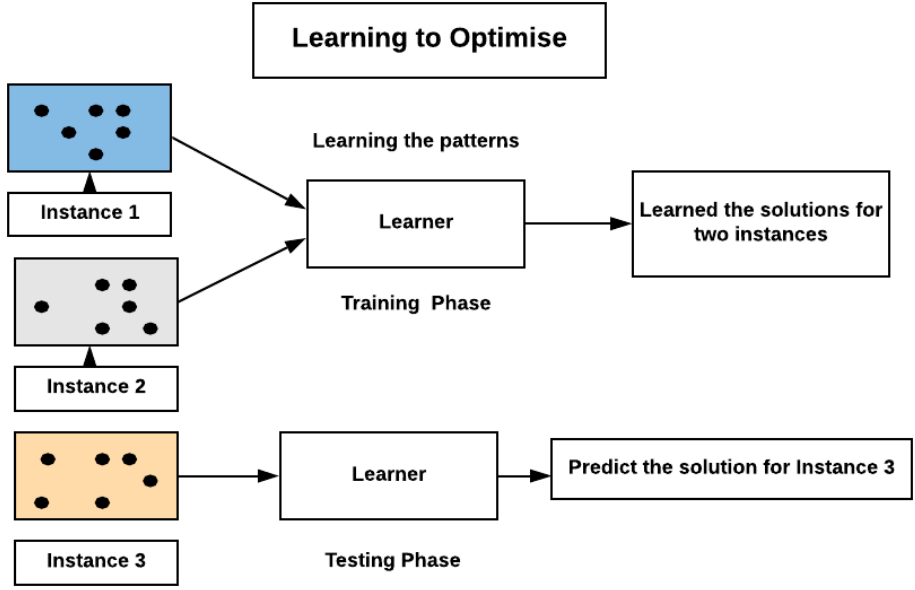}
 \caption{Learning to Optimise }
 \label{fig:optimise}
\end{figure}
\subsection{Measure the Difficulty Level}\label{section:hard}

In \cite{gent1996tsp}, the authors suggested that the hardness of solving the travelling salesman problem is related to how the cities cover the area and spread across the area (space). They analysed how easy and hard a TSP instance to find a new solution with a shorter tour length.  To the effort, authors place this phase transition at $ \sqrt{\frac{l}{N.A}}$ $\approx 0.75$, where N is the number of cities, A the area covered by the cities and l the tour length.  They set a parameter $ \sqrt{\frac{l}{N.A} - 0.75}$  which can be used as hardness indicator, that measure the difficulty to solve a TSP instance. In \cite{cardenas2018creating} they argued that this measurement requires finding the optimal or a quasi-optimal solution of the instance which makes this calculation difficult to use this measure frequently. Therefore their finding is the hardness of TSP instances based on spatial properties of instances such as space and distribution of cities.  In \cite{cardenas2018creating}, given the parameters, A and N, the decision problem becomes more difficult for instances with $ \sqrt{\frac{l}{N.A}}$ $\approx 0.75$. Therefore, the closeness of $\sqrt{\frac{l}{N.A}}$ to 0.75 can provide an additional hardness rank for comparison purposes, and illustrate an example of the most difficult instance Berlin52. 
\begin{figure}
 \centering
  \includegraphics[width=0.85\textwidth]{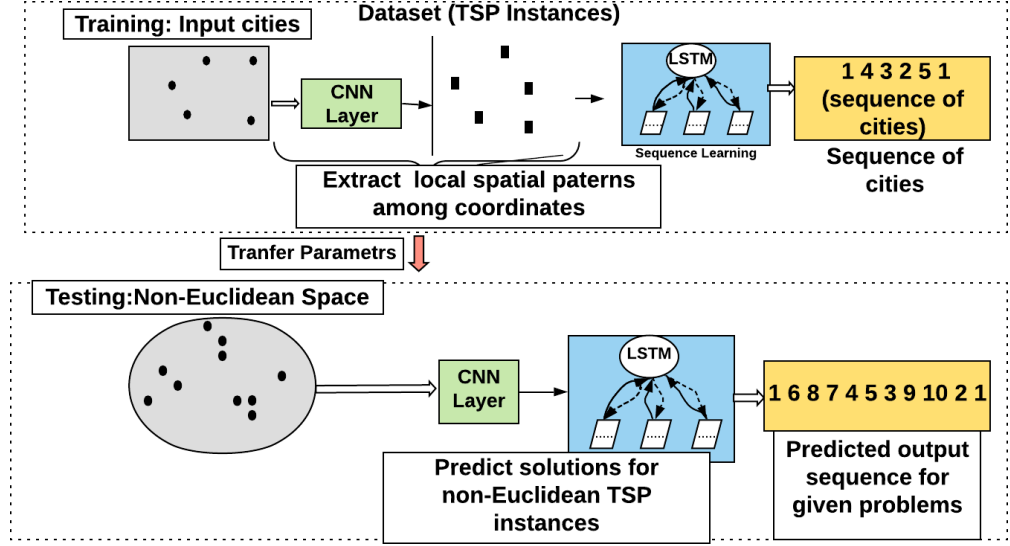}
 \caption{Our proposed model NETSP-Net.}
 \label{fig:NETSP}
\end{figure}

\subsection{Learning to Optimise TSP}\label{section:opt}

In (supervised) learning problems, we generally map inputs to outputs given a training dataset of examples (input, output). (Supervised) Learning to optimise similarly does this mapping and learning, but the inputs are optimisation problems and outputs are their solutions. During training, the learning model learns to map input instances to TSP solutions, then during testing/prediction phase, a solution is predicted for each input instance. This is summarised in Fig.~\ref{fig:optimise}. 

More formally, given a set of training examples $(X_1, Y_1)\cdots(X_n, Y_n)$, a learning algorithm searches for a function, $g: X \rightarrow Y$, where is the optimisation instance/problem space $X$ and the solution space $Y$. Alternatively and equivalently, we want to find a Y solution that minimises the optimality gap. For learning to optimise, we just obtain that mapping function differently to optimisation. 

\section{Proposed Model: NETSP-Net}

Our approach, the Non-Euclidean TSP network (NETSP-net) comprises CNN with two RNN modules with attention mechanism \cite{vinyals2015pointer} The employed encoder and decoder both consists of Long Short-Term Memory (LSTM) cell \cite{hochreiter1997long}. The reason we use LSTM, because, it is difficult for RNNs to handle long-term dependencies in the input sequence, LSTM capable of learning long-term dependencies. We are inspired by previous work \cite{vinyals2015pointer} that used a set of softmax layers, and the overall model is illustrated in Fig.~\ref{fig:NETSP}. This approach allows the model to effectively focus on a specific position in the input sequence rather than predicting an index value. We employ the attention mechanism architecture, depicted in Fig.~\ref{fig:Encoder-Decoder}, where the encoder produces embedding of all input nodes, then fed to the decoder network (green). The decoder produces the sequence S as a solution (TSP tour). Our model uses CNN for the representation of inputs to exploit the spatial invariance of all nodes (cities), as this technique can discover the micro-patterns of TSP instances.  Our particular choice of CNN to generalise to unseen various TSP instances such as Non-Euclidean TSP instances. Thus, 1-Dimensional convolutional layers use as an embedding that maps the inputs into D-dimensional vector space, then the elements are fed as an input to the RNN. The first RNN network reads the embedded inputs and uses another LSTM layer to produce or decode the output of S. In the next section, describe the details of each part.
\subsection{Architectural Details}

\begin{figure}
\begin{center}
 \includegraphics[width=0.85\textwidth]{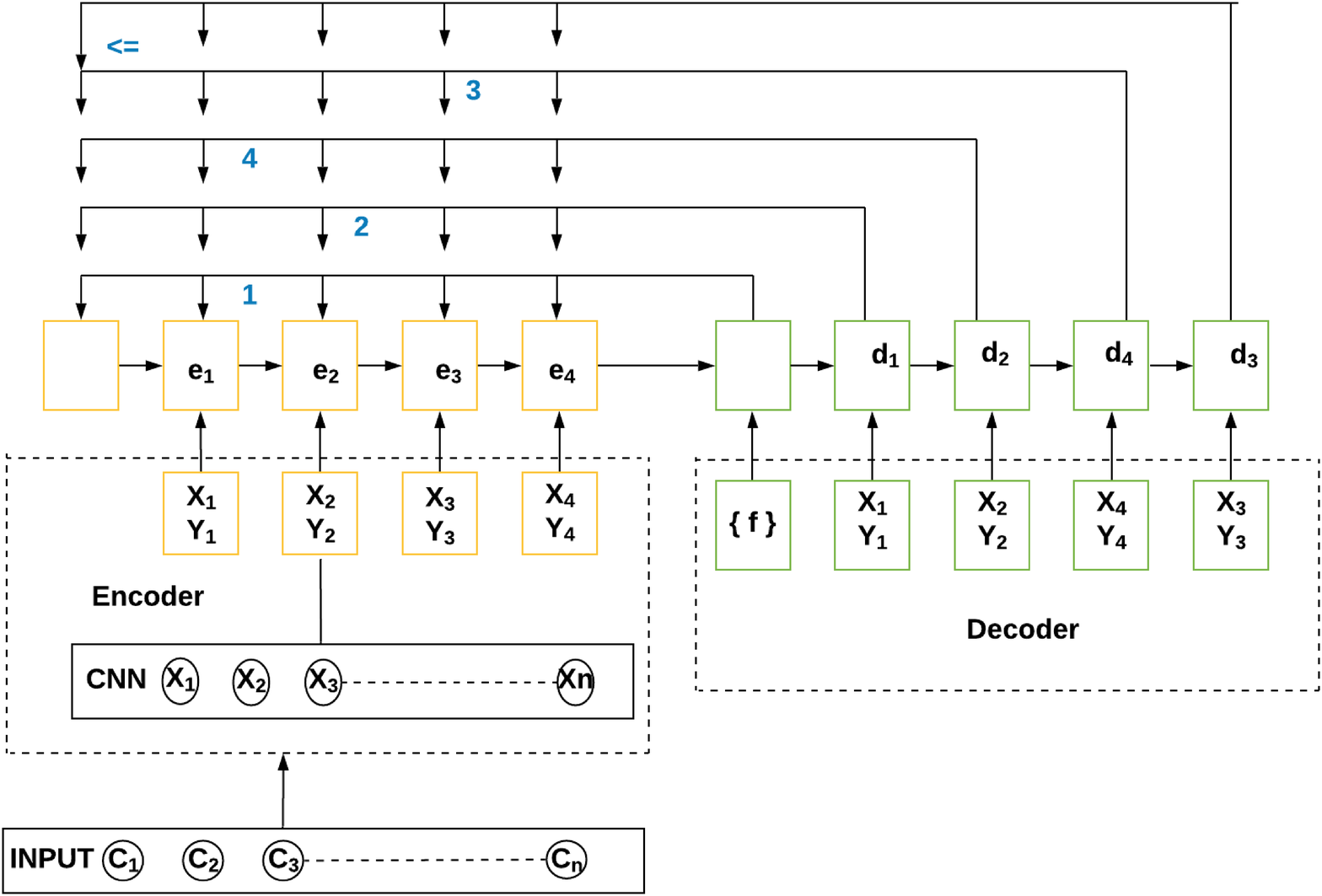}
\caption{NETSP-Net encoder decoder framework. An encoder RNN (yellow) converts the input sequence and fed to the decoder RNN (green) (\cite{vinyals2015pointer}).}\label{fig:Encoder-Decoder}
\end{center}
\end{figure}

\begin{figure}
 \centering
  \includegraphics[width=0.85\textwidth]{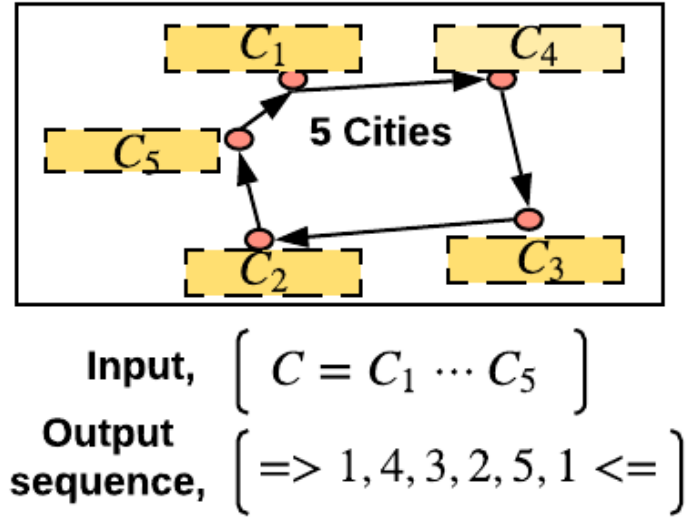}
 \caption{Input/Output representations for TSP. The tokens $=>$ and $<=$ represents beginning and end of sequence respectively.}
 \label{fig:inout}
\end{figure}

In this section, we formally define our NETSP-Net model in terms of TSP. Given a problem instance C, represented as a sequence of n cities in a two-dimensional space, where node $(i\in {1\cdots}n)$. We are interested in finding a minimised tour S, where each city is visited once with a minimum total length. In Fig.~\ref{fig:inout}, for TSP we find an illustration of an input/output pair (C,S). This dominant approach inspired by sequence to sequence model is based on learning general TSP instances, which optimising (training) for the likelihood of the next target city conditioning on the input sequence and the ground truth TSP tour. In the sequence to sequence model, given a training pair, $(C, S)$, it computes the conditional probability $p(S|C;\theta)$, (see in Fig.~\ref{fig:Encoder-Decoder}), estimate the terms of the probability chain rule using a parametric model, i.e.,
\begin{equation}
P(S|C; \theta) = \prod_{i=1}^{m(p)}P_{\theta}(S_i|S_1,\cdots,S_{i-1},C;\theta) 
\label{equation:cross}
\end{equation}
Where $C$ is the sequence of cities, and $S$ is its correct solution.  By maximising the conditional probability for the training dataset, the parameters of the model are learnt, and using the following formulation, where the sum is an over the training example:

\begin{equation}
\theta^* = \arg\max \sum_{C,S} log P(S|C; \theta)
\label{equation:argmax}
\end{equation}

\begin{figure*}
 \centering
  \includegraphics[width=1\linewidth]{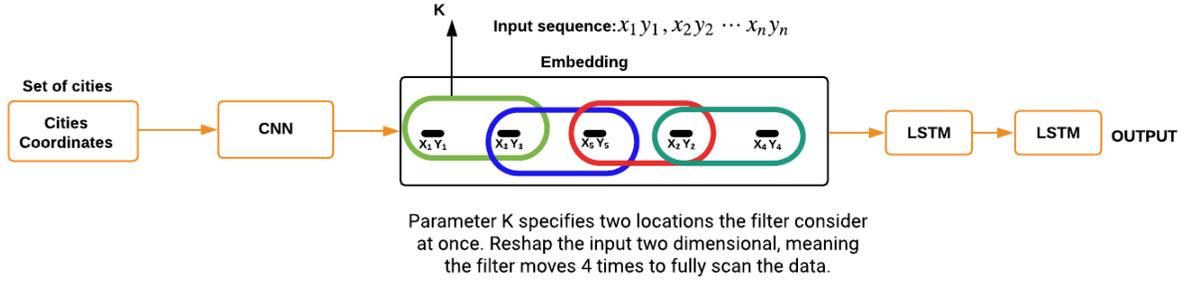}
 \caption{Illustration of the proposed model as applied to a TSP instance. The embedding part illustrates how  Conv1D loop over the input}\label{fig:conv1d}
\end{figure*}

\textbf{Encoder:}
For the encoder, we use 1-dimensional convolutional layers. Unlike languages, when there is no meaningful order in the input, such as TSP, use CNN will be more effective. For example, the inputs are the set of unordered locations (nodes) in the TSP, and any random order holds the same information as the initial inputs. Therefore, in our model, we improve the encoder by directly using the embedded inputs from 1-dimensional convolutional layers. Illustrates in Fig.~\ref{fig:conv1d}, shows while Conv1d scanning over the coordinates of the cities, it can extract the compositional features, for example looping through the existing ones and draw attention to the repeated patterns to discover geometry involved in the instances from the inputs~\footnote{In general, 1D convolutional mechanism, information flows by a convolution operation ($*$) followed by an activation function, $S = f(K*C + B)$, where C and K denotes the incoming input signal and a kernel respectively}. We illustrated in Section~\ref{section:random} to show our claim that embedding layer always gives a better solution (One possible explanation is that the model efficiently extracted the  useful features from the high-dimensional input representations). Here, used embedding layer (CONV1D), in which the in-width is equal to the input length, the number of the filter is F which covers how many different windows from the input is considered. The number of in-channels is the number of elements of C. Input nodes is embedded and processed by CNN layer. From the $d_x$-dimensional input features $x_i$ $(d_x = 2)$, the encoder computes initial $d_h$-dimensional node embeddings (we use $d_h = 128$). Subsequently, the RNN layer in encoder reads the embeddings and generates the latent memory states $(e_i)^n_{i=1}$, where $e_i\in \mathbb{R}^d$. At time-step $i$, the input to the RNN network is a d-dimensional embedding of a 2D point $(X_i)$, which is obtained via a linear transformation of $(X_i)$ shared across all input steps, Fig.~\ref{fig:Encoder-Decoder}. We use a 1-dimensional convolutional network here since the 2-dimensional convolutional does not lead to significant improvement in our experiments. In the next section describes decoder networks. 
 
\textbf{Decoder:}

The decoder network uses an attention mechanism (followed by pointer network attention technique) to produce a distribution over the next city to visit in the tour and generates the latent memory states $(d_i)^n_{i=1}$ where $d_i\in \mathbb{R}^d$ and, at each step i.  Then to the next decoder step decoder receives the next selected city. The first decoder step input (denoted by $\{f\}$ in Fig.~\ref{fig:Encoder-Decoder}) is a d-dimensional vector treated as trainable parameters of our NETSP-Net. During inference, given a sequence C, the learnt parameters select the sequence S with the highest probability of good solutions. (C, S) is the training example pair, and we utilised stochastic gradient descent to optimise the sum of the log probabilities over the training set. Further training details are given in subsection ~\ref{section:5.2}. 

\textbf{Attention mechanism:}
We now describe our attention mechanism. In Fig.~\ref{fig:Encoder-Decoder} demonstrates the attention mechanism \cite{bello2016neural} employed in our model, that, we utilise this to extracts the related information from the inputs. Our attention mechanism, takes input as a query  $q = d_i \in \mathbb{R}^d$  as a vector and a set of reference vectors $R = \{e_1,\cdots , e_k\}$ where $e_i\in \mathbb{R}^d$, and predicts a distribution $A(R, q)$ over the set of k references. This probability distribution represents the extent to which the model points to the reference $r_i$ for the query q. The description of our attention function can be found in Appendix \ref{appendix:AMP}. Some additional computation steps, named glimpses, suggested in \cite{vinyals2015pointer} is used that aggregates the contributions of different parts of the input sequence, like \cite{bahdanau2014neural}. The details of this approach is discussed in Appendix \ref{appendix:AMP}.
\section{Experiments}\label{section:5}

In this section, we discuss the experiment setting for the proposed method. First, the model is evaluated using various TSP instances according to their hardness and use one large instance from TSPLIB library to show that model can generalise on large scale TSP instances in Table \ref{tabile:tsplib}. Second, we consider the non-Euclidean TSP instances, such as Pseudo-Euclidean instances, Geographical (Haversine distance) instances in Fig. \ref{fig:euclidean}. Moreover, we compare the performance of our model on random TSP instances as previous state of the art approaches typically focus TSP random data, reported in Table \ref{tabile:tsp}. We investigated the behaviour of the proposed NETSP-Net approach and compared with other baselines. Our experiments were designed to investigate the following evaluation procedure: 
 \begin{itemize}
    \item We explore the performance of our model on easy, and hard TSP instances. 
    \item In order to verify the generalisation of the proposed method, we test our method on the various edge-weight types of TSP instance.
    \item The performance of our method is evaluated on randomly generated instances that previous work concentrated on.
    \item Analyse model performance using various training and testing size and the result shown.
    \item Another experiment we have done to test the generality of our approach for analysing different set of instances where the set of features use to characterise TSP instances in Appendix ~\ref{appendix:B}. This special feature TSP instances are evaluated on chained Lin-Kernighan(CLK) algorithm in paper~\cite{smith2010understanding}.
     \item Compare the solution quality of 1-dimensional convolutional network with 2-dimensional convolutional networks in Appendix~\ref{appendix:E}. 
\end{itemize} 

In the rest of this section, we discuss our dataset, network settings, baselines we compared and finally present our results.
\subsection{Datasets}

To evaluate our model, we have used the benchmark TSPLIB dataset~\cite{reinelt1991tsplib} (different edge-weight types of TSP instances) and randomly generated instances; dataset was first introduced by~\cite{vinyals2015pointer}. The TSPLIB is a popular benchmark for evaluating TSP algorithms. For the synthetic data, we follow~\cite{vinyals2015pointer} and generate TSP instance by generating points in within a 2D unit square $[0, 1]\times[0, 1]$, uniformly randomly. The test datasets have graphs of sizes 20, 50 and 100 nodes. The training sets consist of one million pairs of problem instances and solutions. The test sets consists of 1000 pairs each.

\subsection{Network Setting}\label{section:5.2}
We used the same architecture settings throughout all the experiments and datasets. In all our  experiments, use mini-batches of 128 sequences, one-dimensional convolutional operation, convolution layer works as an embedding layer which embed each city location is into a vector of size 128, LSTM cells with 128 hidden units and we train our models with the Adam optimiser \cite{kingma2014adam}. For the randomly generated dataset, we use an initial learning rate of 0.001 (considered best learning rate) for TSP20, TSP50, TSP100. The decay rate of every 5000 steps by a factor of 0.96. We clip the L2 norm of our gradients to 1.0. We varied the hyper-parameters and found results are most similar. Run times are important but can vary due to implementation using Python or C++. We used python for implementation. Another important factor is using hardware such as GPUs or CPUs \cite{kool2018attention}. We implemented our experiments on the same hardware platform with Intel Xeon 2.4 GHz with 56 cores. We run Concorde \cite{applegate2006concorde} and OR-Tools \cite{ortools} on Intel $CORE^{TM}$ i5-7200U CPU@2.50 GHz as we do not show run times in our evaluation, used different hardware. 
\subsubsection{Evaluation} 
We report the following metrics to evaluate performance of our model and other baselines. These were previously used in \cite{kool2018attention}, \cite{joshi2019efficient}:
\begin{itemize}
 \item\textbf{Predicted tour length}: Average predicted tour length.
 \item \textbf{Optimality gap}: The average percentage ratio of the predicted tour length relative to optimal solutions \cite{joshi2019efficient}.
\end{itemize}
\subsubsection{Baselines:} \label{section:5.2.2}

\begin{figure}
	\centering
	 \includegraphics[width=.85\linewidth]{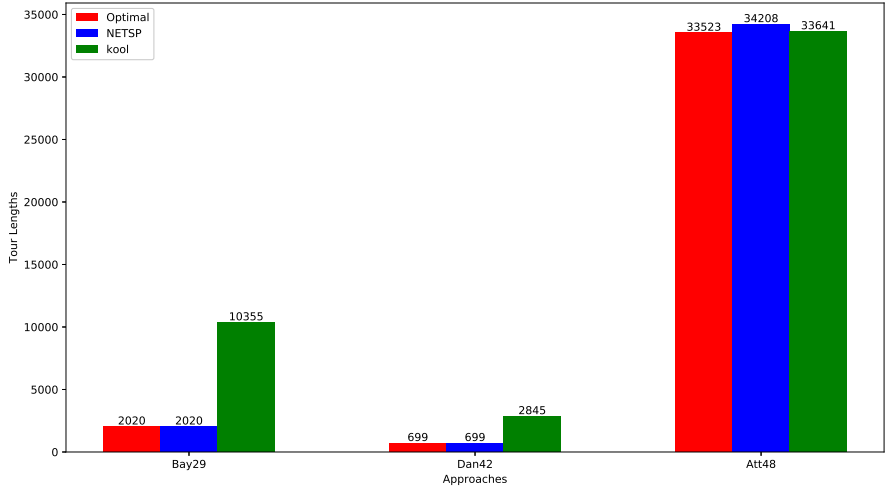}
	\caption{Tour length comparison when solving for various edge-weight types of TSP instances and trained on Euclidean Instances). We only considered~\cite{kool2018attention} algorithm to compare with our approach because~\cite{kool2018attention} outperformed all the previous learning algorithms.}\label{fig:euclidean}
\end{figure}

The performance of our technique compared with a variety of baselines, including: Solvers, an efficient exact solver specialized for TSP; heuristics, well-known heuristic solvers that achieves state-of-the-art performance on various routing problems; open source software for combinatorial optimisation (OR-Tools), a mature and widely used solver for routing problems based on meta-heuristics; learning models using supervised techniques(SL); and learning methods using reinforcement learning (RL). We report optimal results by Concorde specialized for TSP~\cite{applegate2006concorde}. We compare against 2-opt and Christofied local search, MST, Cheapest, Nearest, Random and Farthest Insertion, as well as Nearest Neighbor in Table~\ref{tabile:tsplib} and~\ref{tabile:tsp}. We also focus our comparison to the recently proposed deep learning methods \cite{vinyals2015pointer}, \cite{joshi2019efficient}, \cite{bello2016neural},\cite{deudon2018learning}, \cite{khalil2017learning}, \cite{nazari2018reinforcement} and \cite{kool2018attention} using their publicly released implementations in Table~\ref{tabile:tsp}. We also compared our results with Google Or-Tools \cite{ortools} in both the tables~\ref{tabile:tsplib} and~\ref{tabile:tsp}. The description of baselines, experimental procedures are as follows:
\begin{table*}[h]
\small
\centering
\caption{TSPLIB: Instances are reported by difficulty level (size indicating at the end of the instance name, and easy to hard instances arranged as top to bottom respectively), values (tour length) reported are the cost of the tour found by each method (\textbf{lower is better, best in bold}). Easy and hard instances classified according to \cite{cardenas2018creating} from topmost to bottom-most which means \textbf{Eil51} is the easiest and \textbf{Berlin52} is the hardest. Other methods never tested their model using the TSPLIB library, so we reported results of AM\cite{kool2018attention} using their publicly available code. However, for \textbf{S2V-DQN}~\cite{khalil2017learning} and all heuristics, we reported results from~\cite{khalil2017learning}.}
\begin{tabular}{c|c|c|c|c|c|c|c|c|c|c}\hline
Easy=>Hard & Opt & Furthest & 2opt&Christ& Cheapest &MST & Or& AM& S2V-DQN& NETSP\\\hline
Eil51 & 426 & 567 & 446 & 527 & 494 & 614 & 427& 435 & 439  & \textbf{427}\\\hline
Eil76& 538 & 583 & 591 & 646 & 607 & 743 & \textbf{538}& 564 & 564 & 543\\\hline
Eil101& 629 & 659 & 702 &728  &  693 & 847 & 651& 668& 659 & \textbf{632}\\\hline
St70& 675 & 712 & 753 & 836 & 776 &  858 & \textbf{682}& 829 & 696 & 712\\\hline
Ch150 & 6,528 & 7,210 & 7,232 & 7,641 & 7,995 & 9,203 & \textbf{6632}& 9298 & 6985 & 6870\\\hline
Ch130& 6,110& 6,498 & 6,470 &7,367  & 7,279 & 8,280 & 6147& 6329& 6270 & \textbf{6133}\\\hline
Berlin52& 7,542 & 8,307 & 7,788 & 8,822 & 9,013  &  10,402 & 7542 &7788& 7542 & \textbf{7542}\\\hline
\end{tabular}
\label{tabile:tsplib}
\end{table*}
\begin{figure}
\begin{center}
\includegraphics[scale=.25]{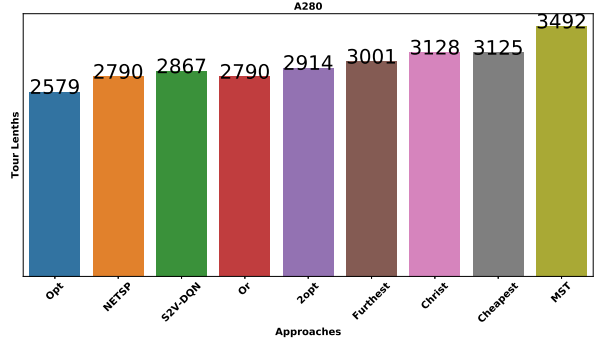}
\caption{Comparison of A280 results: NETSP vs baselines. The best value (tour length) across all methods. The y-axis is the tour lengths compare to the optimal solutions (opt in the figure refer to the optimal solution).}\label{fig:a}
\end{center}
\end{figure}
\begin{itemize}

\item \textbf{Concorde}: Concorde also denoted as (OPT) in Table~\ref{tabile:tsplib} is a computer code for the symmetric TSP and some related network optimisation problems. Concorde's TSP solver has been used to obtain the optimal solutions for all random instances and TSPLIB instances; We implemented Concorde~\cite{applegate2006concorde} that use algorithms~\cite{dantzig1954solution,padberg1991branch,applegate2003implementing}. In Table~\ref{tabile:tsplib} and Fig.~\ref{fig:a} denoted the results as OPT. In Fig.~\ref{fig:euclidean} denoted the result as optimal.

\item\textbf{Nearest Insertion:} Nearest Insertion~\cite{WinNT21} inserts the node to the nearest set of nodes (neighbours), for such insertion operation causes the least cost in the overall tour length.

\item\textbf{Farthest Insertion:} Furthest Insertion~\cite{WinNT21} needs to select two cities and connect them to get the least cost tour, then find another farthest city of this tour. Repeat the step until every city associates to complete the tour

\item\textbf{Random Insertion:} Random Insertion \cite{huang2017investigating} needs to select two cities. A Random insertion adds a random node where the added node order is also random similar to the nearest neighbour.

\item\textbf{Nearest neighbour:} The Nearest Neighbour~\cite{WinNT21} heuristic represents the partial solution as a path with a start and end node. First, start in some city and the select to visit the city to the starting city. Continue the process, and at the end, all cities visited, and the end city is connected with the start city. We follow the implementation of~\cite{kool2018attention}.

\item\textbf{2-Opt:} Croes~\cite{croes1958method} first introduced the 2-optimisation method, which is a simple and very common operator. The idea of 2-opt is to exchange the links between two pairs of subsequent nodes.

\item\textbf{Minimum Spanning Tree:} A Minimum Spanning Tree (MST) \cite{cheriton1976finding} aims to minimise the weights (tour lengths) of the edges of the tree. 

\item\textbf{Christofides:} Christofides algorithm~\cite{christofides1976worst} is a  heuristic algorithm which aims to find a near-optimal solution to the problem. It follows steps, first find an MST (minimum spanning tree ), second find a minimum-weight perfect matching M among vertices with odd degree. Create a minimum spanning tree T of G, to Make a multigraph  G; it combines the edges of M and T. Then find an Euler cycle in G by skipping vertices already visited. 

\item\textbf{Cheapest:} The Cheapest-Link Algorithm select the edge with the smallest weight and mark it and continue that following rules, do not pick an edge that will close a circuit –Do not pick an edge that will create three edges coming out from a single vertex Connect the last two vertices to close the circuit.

\item\textbf{OR-Tools:} Google Optimisation Tools (OR-Tools) is an open-source solver for combinatorial optimisation problems. OR-Tools contains one of the best available vehicle routing problem (VRP), which is a generalisation of the TSP and implemented many heuristics for finding an initial solution and metaheuristics, we use it as our baseline. We have used the local-search meta-heuristics used in OR-Tools as Guided Local Search. 

\item\textbf{Pointer Network(PN)}: We implemented the pointer network with supervised learning \cite{vinyals2015pointer} 

\item\textbf{Bello et al.}: Across all experiments,\cite{bello2016neural} used mini-batches of 128 sequences, LSTM cells with 128 hidden units, and train models with the Adam optimiser \cite{kingma2014adam}. The learning rate used  $10^{-3}$ for TSP20 and TSP50. Larger problems TSP100 learning rate used $10^{-4}$. After implementing the code, we reported the result.

\item\textbf{EAN}: For experiment Policy gradient across all tasks, we follow the configuration provided in (\cite{deudon2018learning}), and we report results.

\item\textbf{[$RL+Gr$ and $RL+Bs$]}: For experiment, Nazari et al.~(\cite{nazari2018reinforcement}) implemented following the GitHub code \footnote{https://github.com/OptMLGroup/VRP-RL}.

\item\textbf{$AM+Gr$}: For experiment,~(\cite{kool2018attention}), initiating parameter uniform $(-\frac{1}{\sqrt{d}}, \frac{1}{\sqrt{d}})$, with $d$ the input dimension. Every epoch we process 2500 batches of 512 instances. Used a constant learning rate 0.0001 Training with a higher learning rate 0.001 is possible and speeds up initial learning, but requires decay (0:96 per epoch) to converge. Implemented following the the GitHub code \footnote{https://github.com/wouterkool/attention-learn-to-route}.
\end{itemize}

\subsection{Results}
In this section, we compare the solutions produced by proposed and baselines methods mentioned in Section~~\ref{section:5.2.2}.
\begin{table*}[h]
\centering
\caption{Average tour length (TourL) and the gap percentages reported with respect to optimal value. * indicate that values are reported from their works. The gap percentages reported with respect to optimal value. In the first column RL means reinforcement learning, SL means supervised learning}
\begin{tabular}{lc|cc|cc|cc}\hline
\multicolumn{2}{c}{Method} & \multicolumn{2}{c}{TSP=20} & \multicolumn{2}{c}{TSP=50} & \multicolumn{2}{c}{TSP=100} \\ \hline
\multicolumn{2}{c|}{Solver}  & TourL  & Gap & TourL  &  Gap & TourL   & Gap  \\\hline\hline 
&Concorde      & 3.84  & 0.00\%  & 5.70  & 0.00\% & 7.77  & 0.00\%\\ \hline
&\textbf{Heuristics} &  &  & &  &  &    \\\hline
&Nearest Neighbour &  4.50  & 17.18\%  & 7.00 & 22.80\% & 9.68 &24.58\%\\ 
&Nearest Insertion &  4.33  & 12.76\%  & 6.78 & 18.94\% &  9.45 &21.62\%\\ 
&Random Insertion & 4.00   & 4.16\%  & 6.13 & 7.54\% & 8.51 &9.52\%\\ 
&Farthest Insertion & 3.92  & 2.08\%  & 6.01 & 5.43\% & 8.35  &7.46\%\\ \hline
&\textbf{Meta-heuristic} &  &  & &  &  &    \\\hline
&{Or-tools} &  3.85 & 0.26\% & 5.80  & 1.75\% & 8.30  & 2.90\% \\\hline
&\textbf{Learning Models (SL)} &  &  & &  &  &     \\\hline
&PN \cite{vinyals2015pointer}  & 3.88  & 1.03\% & 6.62 & 16.14\%  &  10.88 & 40.20\%\\
&GCN* \cite{joshi2019efficient}& 3.86  & 0.52\% & 5.87   & 2.98\% & 8.41 & 8.23\%\\\hline
&\textbf{Learning Models (RL)} &  &  & &  &  &    \\\hline
&Bello et al.\cite{bello2016neural} & 3.89 &  1.30\% & 5.99  & 5.08\% & 9.68  &24.73\%\\
&EAN. \cite{deudon2018learning} & 3.93  & 2.34\%  & 6.63   & 16.31\% & 9.97 & 28.31\%\\
&S2vDQN* \cite{khalil2017learning} & 3.89  & 0.26\% & 5.99 & 1.75\% &\textbf{8.31} & 7.07\%\\
&RL+Gr\cite{nazari2018reinforcement} & 4.00  & 4.16\% & 7.01 & 22.98\% & 9.46 &21.75\%\\
&RL+BS\cite{nazari2018reinforcement} & 3.96  & 3.12\% & 7.40 &  28.82\% & 8.93  &14.92\%\\
&AM+Gr\cite{kool2018attention}& 3.87 &0.78\%  & \textbf{5.80}&1.75\%   & \textbf{8.15}  &4.89\%\\\hline
&\textbf{NETSP-Net (Ours)} & \textbf{3.85}  & 0.26\%  & {5.85}  & 2.63\% & {8.31}  & 6.94\% \\\hline
\end{tabular}
\label{tabile:tsp}
\end{table*}
\subsubsection{Generalisation on Different Distributions and Larger Instances} \label{section:various} 
In this section, our target is to investigate our model's performance on different distribution. We use the standard TSPLIB library \cite{reinelt1991tsplib}, which is publicly available. We show that the model generalises reasonably well to large problems and a very different distribution. Moreover, the model was trained on instances with no more than $n=50$ cities. We classify TSP instances by how they are distributed on the Euclidean plane, which indicates the hardness of the instances. Note that these instances may follow distributions that are completely different from those used in training, e.g., in node location patterns.  In Table~\ref{tabile:tsplib}, we have reported the optimal TSP solutions implemented using Concorde \cite{applegate2006traveling}, then compare against other learning methods AM, \cite{kool2018attention} (using their publicly available code using benchmark TSPLIB dataset, trained using random data) and S2V-DQN \cite{khalil2017learning} (reported their result). Furthermore, compare against other heuristics, e.g., Furthest, 2-opt, Christofides, Cheapest, MST and OR-tools \cite{ortools}. Also, we arrange the instances from easy to hard (in the first row is the easy instance, and the last row is the hardest instances). It is clear that the structure of TSP instances influences the solution quality, which our model able to understand and shows that our approach performs relatively well on these instances. Table \ref{tabile:tsplib} demonstrates that our model performs well on the various hardness of TSP instances when compared to other two learning algorithms \cite{khalil2017learning} and \cite{kool2018attention}, which means the model can generalise beyond random data (benchmark data). We show that the quality of our solution does not degrade very fast with the increase of problem size, but other approaches lost performance progressively faster for larger problem sizes in Table~\ref{tabile:tsplib} and in Fig.~\ref{fig:a}. 
For larger problem size, we wanted to know, what extent the learned algorithm generalises to larger problem sizes. In Fig.~\ref{fig:a}, we illustrate the most larger problems A280 data (from TSPLIB data) learned by our model. It compares the gap to the optimal solution, with a smaller gap been more desirable. The optimal solutions (opt) are obtained from the Concorde algorithm. We observed that for large-scale TSP instances, the NETSP-Net outperforms both state-of-the-art learning approaches~\cite{khalil2017learning} and~\cite{kool2018attention}, which demonstrates the usefulness of our approach. The convolutional neural network has shown to be powerful to represent the spatial patterns among coordinates. Embedding with a convolutional network increases the learning efficiency and generalises to larger problems resulting in predicted solutions for various TSP instances. 
%
\begin{table*}[h]
\small
\centering
\caption{In this table, we compared our result with the most recent work (\cite{kool2018attention}), train with various sizes of instances and test on various sizes of instances. Average tour length (TourL), and the gap percentages reported with respect to optimal value for TSP20 3.84, TSP50 5.70 and TSP100 7.77 using Concorde (\cite{applegate2006concorde})}
\begin{tabular}{c|cc|cc|cc}
\hline
Method & \multicolumn{2}{c}{Training  Size} & \multicolumn{2}{c}{Testing  Size}& \multicolumn{2}{c}{TourL} \\ \hline
\textbf{}    & Problem  & Size & Problem  &  Size & TourL   & Gap\\\hline 
\textbf{Learning Models (RL)} &  &  & &  &  &    \\\hline
AM.\cite{kool2018attention} & TSP  & 20 & TSP & 20 & 3.85 & 0.26\% \\ \hline
AM.\cite{kool2018attention} & TSP  & 20 & TSP & 50 & 5.95 & 4.38\%\\ \hline
AM.\cite{kool2018attention} &  TSP  & 20 & TSP & 100 & {9.18}& 18.14\%\\\hline
AM.\cite{kool2018attention}   & TSP  & 50 & TSP & 20& 3.87 &0.78\%  \\ \hline
AM.\cite{kool2018attention}    & TSP  & 50 & TSP & 50& \textbf{5.80} & 1.75\% \\ \hline
AM.\cite{kool2018attention} & TSP  & 50 & TSP & 100 & \textbf{8.15} & 4.89\% \\ \hline
AM.\cite{kool2018attention}   & TSP  & 100 & TSP & 20& 4.14& 7.81\% \\ \hline
AM.\cite{kool2018attention}    & TSP  & 100 & TSP & 50& 5.94& 4.21\% \\ \hline
AM.\cite{kool2018attention}    & TSP  & 100 & TSP & 100& \textbf{8.12}& 4.50\% \\ \hline
\textbf{Learning Models (Ours)} &  &  & &  &  &     \\\hline
{NETSP-Net}   & TSP  & 20 & TSP & 20 & 3.85 & 0.26\% \\ \hline
{NETSP-Net}  & TSP  & 20 & TSP & 50 & \textbf{5.81} & 1.92\%\\ \hline
{NETSP-Net} &  TSP  & 20 & TSP & 100 & \textbf{9.01}& 15.95\%\\\hline
{NETSP-Net}   & TSP  & 50 & TSP & 20& \textbf{3.85} &0.26\%  \\ \hline
{NETSP-Net}   & TSP  & 50 & TSP & 50& 5.85 & 2.63\% \\ \hline
{NETSP-Net} & TSP  & 50 & TSP & 100 & 8.31 & 6.94\% \\ \hline
{NETSP-Net}    & TSP  & 100 & TSP & 20& \textbf{3.86}& 0.52\% \\ \hline
{NETSP-Net}   & TSP  & 100 & TSP & 50& \textbf{5.87}& 2.98\% \\ \hline
{NETSP-Net}  & TSP  & 100 & TSP & 100& 8.30& 2.90\% \\ \hline
\end{tabular}
\label{tabile:size}
\end{table*}
\subsubsection{Generalisation on NON-Euclidean TSP Instances}
In this analysis, we show the results of evaluating the different edge-weight types from the TSPLIB library. In the TSPLIB library, very few instances we have with other edge-weight. Previous learning approaches never evaluate their model with various edge weight types of instances. In Fig.~\ref{fig:euclidean}, we show the performance on various edge-weight types of TSP instances. Compared with the optimal solutions provided by Concorde \cite{applegate2006concorde} and one learning to optimise method \cite{kool2018attention}. The Fig.~\ref{fig:euclidean}, illustrates that NETSP-Net model can generalise beyond Euclidean instances. In Fig.~\ref{fig:euclidean} Bays29 instance has its distance matrix computed according to Haversine formula (great circle distance). We reason this performance boost is caused by the use of a CNN, which understands the patterns of TSP instances and predicts the solution disregarding distance.
\subsubsection{Result Analysis on TSP Random Data}\label{section:random}

In this section, we want to evaluate and show our performance is comparable with existing work as the previous state of the art approaches typically focus TSP random data. For the TSP, we report optimal results by Concorde \cite{applegate2006traveling} and \cite{helsgaun2000effective}. Besides, we compare against Nearest, Random and further Insertion, as well as Nearest Neighbour. Table~\ref{tabile:tsp} is separated into four sections: solver; heuristics; learning methods using reinforcement learning (RL), and; learning models using supervised techniques (S). For all results were implemented using their publicly available code except Graph Convolutional Network (GCN )\cite{joshi2019efficient} taken from  their paper. Likewise, all other methods we reported tour length. We implemented  PN., \cite{vinyals2015pointer}, Bello et al., \cite{bello2016neural}, EAN., \cite{deudon2018learning}, Kool et al., \cite{kool2018attention}, \cite{nazari2018reinforcement} and accordingly refer to the results we found from our implementation. We achieved comparable results to the best solver~\cite{kool2018attention} and reported the average tour lengths of our approaches on TSP20, TSP50, and TSP100 in Table~\ref{tabile:tsp}. Additionally, we compared against another two methods and reported their results \cite{khalil2017learning} and \cite{xing2020graph}. In this experiment, the average tour length, for instance, TSP50 and TSP100 \cite{kool2018attention} performed better than our approach, so we have reported a statistical analysis test to analyse the two groups of result in Appendix~\ref{appendix:D} and there is no significant difference between two sets of the result. 

\subsection{Result Analysis on Various Training and Test sizes}\label{appendix:A}
In Table \ref{tabile:size}, we evaluate the effect of various training and test sizes to understand the better impact of the learning paradigm. Our investigation focuses on various TSP sizes as training data and test data. We aim to analyse the behaviour of one state-of-the-art learning model and compared with our work. We have trained each approach with TSP20, TSP50 and TSP100 instances and tests with various instances to analyse the behaviour of approaches in Table \ref{tabile:size}. AM \cite{kool2018attention} generates better results for TSP100 size of data when trained the model with both 50 and 100 nodes. From this study, we can conclude when we combine a different set of training, and testing sizes our model outperformed~\cite{kool2018attention}, for most of the set of training and test pairs which implies even have a small number of a dataset for training the model, our NETSP-Net able to generate feasible solutions for a given TSP instance. 
\section{Conclusion}
We design a neural architecture based on a convolutional neural network combined with a Recurrent Neural Network (RNN) to enable learning general TSP problems.  This paper demonstrates that by combining CNN with LSTM layers, we able to increase the learning efficiency and subsequently able to learn new problem instances with different data sizes, settings, and types. Our NETSP-Net can learn clever heuristics (or distributions over city permutations) for the TSP, which CNN helps to learn local patterns among coordinates that helps the model to generalise to various TSP instances. The model trained with a considerable number of TSP problems and their solutions, also empirically demonstrates that the model trained once on small data performed well on larger data. The goal of this work is not to outperform the existing state of the art TSP learning algorithm. Here, our research is to show the direction to learn general TSP instances in terms of the difficulty level of instances and edge-weight types and problem sizes. The model also gives reasonably good solutions on benchmark datasets. Empirically, our method outperforms state-of-the-art deep models on both various TSP instances. We want to note that our method has great potentials in learning a variety of more complex types of combinatorial optimisation problems, e.g. scheduling and future plan to investigate these possibilities. Our future work would be to improve the model to further scaling to larger problems and real-world problems. We believe that our method is an important beginning for generalising learning heuristics for all types of TSP instances.

\section*{Acknowledgements}
The authors wish to thank Kendall Taylor for his valuable comments and helpful suggestions for figures which greatly improved the paper’s quality.
\bibliographystyle{unsrt} 
 \bibliography{my}
\newpage

\appendix
\section{Pointer and Attention Mechanism}\label{appendix:AMP}
The attention mechanism used following pointer network \cite{vinyals2015pointer}. The computations of the attention mechanism are parameterised by  $\alpha_R$ , $\alpha_q \in \mathbb{R}^{d\times d}$ attention metrices and an attention vector $v \in \mathbb{R}^d$ as follows:
\begin{equation}
  u_{i} =
    \begin{cases}
       v^T tanh(\alpha_R.r_i + \alpha_q.q) \\
       \text{if   $i\neq s(j)$  for all $j<i$  for $i=1,2,\cdots k$} \\
      
    \end{cases}   
    \label{equation:attention}
\end{equation}
\begin{equation}
A(R, q;\alpha_R,\alpha_q, v) \overset{def}{=}  softmax(u_i).
\end{equation}
Here softmax normalise the vector $u_i$ and the normalise vectors distribution over the inputs. The output of our model are the parameters $v$, $\alpha_R$ , $\alpha_q$  which are learnable. The function of the pointer mechanism is to select the probability of visiting the next node s(j)  of the tour  at decoder step j as follows:
\begin{equation}
p(s(j)|s(<j), C) \overset{def}{=}  A(e_{1:n}, d_j).
\end{equation}
As shown in Equation ~\ref{equation:attention}, ensures that our model only points at cities using $u_i$ as pointer to the input elements and hence outputs valid TSP tours S.
We use glimpse function in the attention mechanism. Our glimpse function F(R, q), receives the same inputs as the attention function A, and its computations are parameterised by $\alpha_{R}^f,\alpha_{q}^f \in \mathbb{R}^d \times d$ and $v^f \in \mathbb{R}^d$. It performs the following equation.

\begin{equation}
p = A(R, q;\alpha_R^f,\alpha_q^f, v^f) 
\end{equation}

\begin{equation}
F(R, q;\alpha_R^f,\alpha_q^f, v^f) \overset{def}{=} \sum_{i=1}^k r_ip_i
\end{equation}
The glimpse function F computes a linear combination of the reference vectors.  The attention probabilities weighted the reference vector. The same reference set R can be used many times:
\begin{equation}
f_0 \overset{def}{=} q
\end{equation}

\begin{equation}
f_l \overset{def}{=} F(R, f_{l-1};\alpha_R^f,\alpha_q^f, v^f)
\end{equation}
Finally, the most favourable outcome $f_l$ vector is passed to the attention function $A(R, f_l; \alpha_R ,\alpha_q, v)$, and generate the probabilities of the pointing mechanism.

\section{Result Analysis on Various difficulty level of Training and Test TSP Instances}\label{appendix:B}
In  Table \ref{tabile:net}, we analyse to understand the performance of our approach on various TSP difficulty level of instances \cite{smith2010understanding}. In \cite{smith2010understanding}, an evolutionary algorithm on instances of the TSP are are used to produce different classes of instances that are easy or hard for certain algorithms.  In this paper, they analysed an extensive set of features to characterise instances of the TSP and studied their impact on the difficulty for each algorithm. We evaluated our approach with instances that are easy or hard for Chained Lin-Kernighan (CLK) algorithm described in \cite{smith2010understanding}. In this experiment, we first trained our model with easy instances and tested on easy and hard instances, then we trained our model with hard instances and evaluated our model with easy and hard instances \cite{smith2010understanding}. We have done the same experiment for \cite{kool2018attention} and illustrates the result in Table \ref{tabile:kool}. From this experiment, we can conclude our model outperformed Kool et al. \cite{kool2018attention} of the instances analysed in \cite{smith2010understanding}.
\begin{table*}[h]
    \caption{In Smith et al.~\cite{smith2010understanding}, analysed various algorithms on special characteristics TSP instances, we explore our approach and compared with \cite{kool2018attention} on TSP instances reported in the paper~\cite{smith2010understanding}. Trained NETSP-Net and Kool et al \cite{kool2018attention} with easy and hard instances, test on easy and hard instances. Optimal value for all instances reported using Concorde \cite{applegate2006concorde}}
     \centering
     \begin{subtable}{0.45\linewidth}
    \begin{minipage}{0.45\linewidth}
      \caption{NETSP-Net}
\begin{tabular}{c|c|c|c|c}         \hline
Problems & \multicolumn{3}{c}{Testing} \\ \hline
Training    & Easy  &    Hard  & Concorde\\\hline
Easy TSP50     & 2872  & 3114 & 2823\\  \hline
Hard TSP100     & 2865 & 3002 & 3031\\  \hline  
        \end{tabular}
        \label{tabile:net}
    \end{minipage}%
    \end{subtable}%
     \begin{subtable}{.45\linewidth}
    \begin{minipage}{.45\linewidth}
      \centering
        \caption{Kool et al.\cite{kool2018attention}}
        \begin{tabular}{c|c|c|c|c}
            \hline
Problems & \multicolumn{3}{c}{Testing} \\ 
\hline
Training    & Easy  &    Hard  & Concorde\\\hline
Easy TSP50     & 9988  &  10934 & 2823\\  \hline
Hard TSP100     & 10937 & 9979  & 3031 \\  \hline
        \end{tabular}
        \label{tabile:kool}
    \end{minipage} 
    \end{subtable}%
\end{table*}
\section{Sensitivity Analysis} \label{appendix:D}

We run sensitivity analysis tests to evaluate the effect of the hyper-parameters used in our approach. There are many parameters in NETSP-Net, and for these tests, we consider the number of hidden units and one best learning rate. We denote $h_1$ as a network with 128 hidden units and $h_2$ as a network containing 256 hidden units. A learning rate of 0.001 is denoted by $l_r1$. For all problems we consider the hyperparameter combination of $h_1$ and $l_r1$ and compare it with $h_2$ and $l_r1$. At the 0.05 significance level, results show that there was no significant difference when the number of hidden units was increased from 128 to 256. Sensitivity Analysis Test to evaluate the effect of Hyper-parameter of our Approach for TSP20, TSP50 AND TSP100 (Parameters($h_1+l_r1$ compared with $h_2+l_r1$) is 0.593898, 0.653731, 0.823725 respectively.

Latter, we run sensitivity analysis tests to evaluate the performance of our approach for instances TSP50 and TSP100. The T-test P-values at the 0.05 significance level for NETSP-Net is compared to  \cite{kool2018attention}, as their model performs better compared to ours for TSP50 AND TSP100 instances. We compared against \cite{kool2018attention} to check there is no significant difference between two sets of data ( 0.4111224  and  0.607879 for TSP50 and TSP100) at the 95\% confidence interval.


\begin{figure*}
 \centering
  \includegraphics[width=1\linewidth]{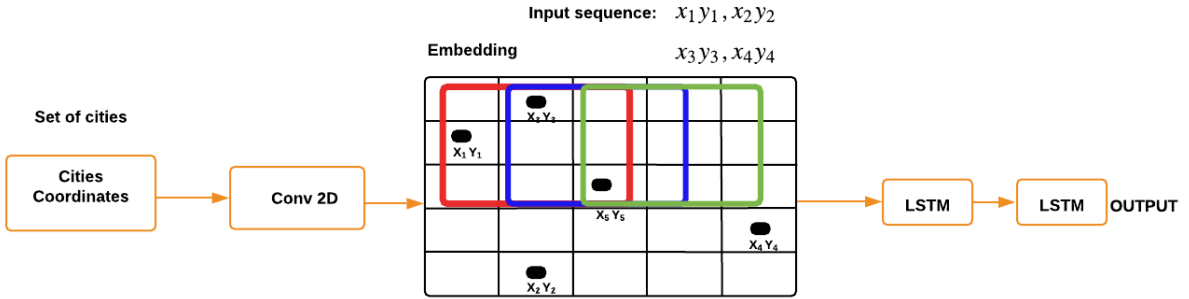}
 \caption{Illustration of proposed model as applied to a TSP instance. The embedding part illustrates how  Conv2D loop over the input}\label{fig:Representation}
\end{figure*}
\section{NETSP-Net(CONV2D)}\label{appendix:E}

We find that training with an embedding layer always yields a quality solution. Additionally, we have evaluated with 2D CNN, and it produces similar results to the 1D. Since 1D is faster to compute, we use 1D CNN. Our proposed framework for a problem with a given set of inputs, we present each input by a sequence of tuples. One can view a vector of features that describes the state of the input. For instance, in the TSP, the convolution operation loops through the whole sequence to learn the pattern of coordinates of the cities.  Looping through the sequence gives a snapshot of the locations, the 2-dimensional coordinates of the location using CNNs for the embedding. In this experiment, we use 2-dimensional convolutional neural networks for the embedding (rather than 1-dimensional convolution layers), and the number of filters $F$. We aim to illustrate the comparison performance 1D CNN to 2D CNN. In Fig.~\ref{fig:Representation}, we illustrate the representation of inputs into a D-dimensional vector space. We demonstrate the solution quality for 2-dimensional convolution layers and 1-dimensional convolution layers for TSP instances on our dataset. As we have coordinates of cities, a 2-dimensional convolutional network convolves with feature parameters across space and can access the spatial information between two coordinates (like in  Conv1D). Using the  Conv2D model does not give us much benefit in terms of tour length. However, our assumption is, solving problems (weighted graphs) such as TSP with time windows, capacitative Vehicle Routing Problems, and an end to end learning problems such as similarity measure of assignment problems \cite{zhang2019end}, our proposed model (with  Conv2D) perform well. Table \ref{tabile:data}, shows the solution quality of 2-dimensional convolution layers and 1-dimensional convolution layers on synthetic data that demonstrates using 2D CNN instead of 1D CNN not obtained much benefit in terms of the solution quality.
\begin{table}[htp]
\small
\centering
\caption{Average Tour length.}
\begin{tabular}{|l|l|l|l|l|}
\hline
Problems & TSP20 & TSP50 & TSP 100 \\\hline
 Conv1D & 3.85 & 5.85 & 8.31 \\ \hline 
 Conv2D & 3.96 & 6.06 & 8.44  \\\hline
\end{tabular}
\label{tabile:data}
\end{table}
\end{document}